\documentclass[letterpaper, 10 pt, conference]{ieeeconf}  

\IEEEoverridecommandlockouts                              

\usepackage{mathptmx} 
\usepackage{times} 
\usepackage{amsmath} 
\usepackage{amssymb}  

\usepackage{cite}
\usepackage{graphicx} 
\usepackage{float}
\usepackage{bm}
\usepackage{color}

\title{\LARGE \bf
	Ergodic Coverage In Constrained Environments Using Stochastic Trajectory Optimization
}

\author{Elif Ayvali, Hadi Salman and Howie Choset
\thanks{*This work was 	supported by NRI IIS-1426655 and ONR N00024-13-D-6400 subcontract through APL/JHU.}
\thanks{$^{1}$E. Ayvali, H.Salman, H. Choset are with the Robotics Institute at Carnegie Mellon University, Pittsburgh,PA 15213, USA
	{\tt\small (eayvali@,hadis@andrew,choset@) cmu.edu}}
}

\begin{document}

\maketitle
\thispagestyle{empty}
\pagestyle{empty}

\begin{abstract}

In search and surveillance applications in robotics, it is intuitive to spatially distribute robot trajectories with respect to the probability of locating targets in the domain. Ergodic coverage is one such approach to trajectory planning in which a robot is directed such that the percentage of time spent in a region is in proportion to the probability of locating targets in that region. In this work, we extend the ergodic coverage algorithm to robots operating in constrained environments and present a formulation that can capture sensor footprint and avoid obstacles and restricted areas in the domain. We demonstrate that our formulation easily extends to coordination of multiple robots equipped with different sensing capabilities to perform ergodic coverage of a domain.




\end{abstract}

\section{Introduction}
Many applications in robotics require efficient strategies to explore a domain for surveillance~\cite{caffarelli2003stochastic}, target localization~\cite{murphy2008search} and estimation tasks~\cite{dunbabin2012robots,smith2011persistent}. Directing a robot to exhaustively pass over all points in the exploration domain may be time consuming or impractical when there is limited power budget. Therefore, probabilistic approaches were developed to exploit prior information, when it exists, to direct the exploration to take less time~\cite{thrun2005probabilistic,Acar_2003_6159,bourgault2003optimal}.
Some of these approaches incorporate a coverage metric, called ergodicity, as an objective to guide the exploration with respect to a desired spatial distribution of robot trajectories\cite{mathew2011metrics,miller2013}.
For example, given a probability density function (pdf) defined over the exploration domain, such as one indicating where survivors might be located in a search and rescue scenario, an ergodic coverage strategy directs the robot to distribute its time searching regions of the domain in proportion to the  probability of locating survivors in those regions. 

The ergodic coverage algorithm, originally formulated by Mathew and Mezi{\'c} ~\cite{mathew2011metrics}, applies to simple kinematic systems and has a natural extension to centralized multi-robot systems. The formulation, however, treats a robot as point mass, and robot's sensor footprint is represented as a Dirac delta function for computational tractability. Additionally, previous implementations of the algorithm in ~\cite{mathew2011metrics,miller2013} do not consider the obstacles in the exploration domain.  To exclude the coverage of certain areas, Mathew and Mezic~\cite{mathew2011metrics} set the value of the desired spatial distribution of trajectories to zero in those areas. However, the robots still visit those areas although the percentage of time spent is small. We have recently employed a potential field based-approach to extend the formulation in ~\cite{mathew2011metrics}, which considers  first-order and second-order systems, to avoid circular obstacles in a planar domain~\cite{Salman2017}. However, generalization of the derivations for other  systems and types of obstacles are nontrivial. Constraints corresponding to arbitrary obstacles can be non-smooth, and, as a consequence, gradient-based optimization methods employed in aforementioned works require special differentiation to guarantee convergence of the solution~\cite{clarke2008nonsmooth}.

In this work, we incorporate the metric for ergodicity~\cite{mathew2011metrics}, as an objective in a sampling-based stochastic trajectory optimization framework.
The idea behind this framework is to construct a probability distribution over the set of feasible paths and find the optimal trajectory  using the cross entropy method~\cite{KobilarovRSS,rubinstein2013cross}. We chose this approach because it generalizes to nonlinear robotic systems operating in constrained environments and can be coupled with other commonly used sampling-based planners such as rapidly-exploring random trees (RRT) and its variants~\cite{karaman2011anytime} to generate the set of feasible paths~\cite{kobilarov2012cross}.

The novelty of the work in this paper lies in the construction of a new ergodic coverage objective that can take into account the robots' sensor footprint, and in the formulation of the algorithm within a stochastic trajectory optimization framework that lends itself to constraints rising from arbitrary-shaped obstacles. Our formulation also allows a robot with a wide sensor footprint to perform coarse ergodic coverage of the domain, while a robot with narrow sensor footprint performs dense ergodic coverage of the same domain.\footnote{Source codes are publicly available at https://github.com/biorobotics/stoec}

This paper is organized as follows. Section~\ref{sec:Background} provides background on ergodic coverage and stochastic trajectory optimization using the cross entropy method. In Section ~\ref{sec:Stochastic_Ergodic}, we derive means to incorporate sensor footprint and introduce a new ergodic coverage objective. In Section ~\ref{sec:Results}, we provide a comparison with the formulation in~\cite{mathew2011metrics}. We also demonstrate the flexibility of our formulation in encoding additional tasks. We present an example in which a robot is directed to perform ergodic coverage while having to reach a destination in finite time.

\section{Background}
\label{sec:Background}

\subsection{Ergodic Coverage}
Ergodic theory is the statistical study of time-averaged behavior of dynamic systems~\cite{petersen1989ergodic}. A system exhibits ergodic dynamics if it uniformly explores all its possible states over time.
Mathew and Mezi{\'c}~\cite{mathew2011metrics} introduced a metric to quantify the ergodicity of a robot's trajectory with respect to a given probability density function. 

Formally, the time-average statistics of a robot's trajectory, \mbox{$\gamma:(0,t]\rightarrow X$}, quantifies the fraction of time spent at a point, $\bm{x}\in X$, where $X\subset \mathbb{R}^d$ is a $d$-dimensional  domain. 
Mathew and Mezi{\'c}~\cite{mathew2011metrics} define the time-average statistics of a trajectory at a point $\bm{x}$ as
\begin{equation}
\Gamma(\bm{x})=\frac{1}{t}\int_{0}^{t} \delta(\bm{x}-\gamma(\tau)) d\tau ,
\label{eq:time-average}
\end{equation}
where $\delta$ is the Dirac delta function.

Let  $\xi(\bm{x})$ be a pdf -- also referred as the desired coverage distribution -- defined over the domain. The ergodicity of a robot's trajectory with respect to $\xi(\bm{x})$ is defined as
\begin{equation}
\Phi(t)=\sum_{k=0}^{m} \lambda_k \left| \Gamma_k(t)-\xi_k\right|^2, \\
\label{eq:metric_ergodicity}
\end{equation}
where \mbox{$\lambda_k=\frac{1}{(1+|k|)^s}$} is a coefficient that places higher weights on the lower frequency components and \mbox{$s=\frac{d+1}{2}$}. The  $\Gamma_k(t)$ and $\xi_k$ are the Fourier coefficients of the distributions  $ \Gamma(\bm{x})$ and $\xi(\bm{x})$ respectively, i.e.\footnote{If there are $N$ robots, the time average statistics and the corresponding Fourier coefficients are defined as \mbox{ $\Gamma(\bm{x})=\frac{1}{Nt}\sum_{j=1}^{N} \int_{0}^{t} \delta(\bm{x}-\gamma_j(\tau)) d\tau $} and \mbox{ $\Gamma_k(t) = \frac{1}{Nt}\int_{0}^{t} \sum_{j=1}^{N}f_k(\gamma_j(\tau))d\tau$}, respectively.},

\begin{equation}
\label{eq:Fourier_coefficients}
\begin{gathered}
\Gamma_k(t) = \langle \Gamma, f_k \rangle = \frac{1}{t}\int_{0}^{t}{f_k(\gamma(\tau))d\tau} \\
\xi_k = \langle \xi, f_k \rangle= \int_{X}\xi (\mathbf{x})f_k(\mathbf{x})d\mathbf{x}
\end{gathered}
\end{equation} 
where $f_k(\bm{x}) = \frac{1}{h_k}\prod_{i=1}^{m}{\cos(\frac{k_i\pi}{L_i}x_i)} $ is the Fourier basis functions that satisfy Neumann boundary conditions on the domain $X$, $m \in \mathbb{Z}$ is the number of basis functions, and $\langle \cdot, \cdot \rangle$ is the inner product with respect to the Lebesgue measure. The term $h_k$ is a normalizing factor. 

The goal of ergodic coverage is to generate optimal controls $\bm{u}^*(t)$ for a robot, whose dynamics is described by a function $\dot{\bm{q}}(t)=g(\bm{q}(t),\bm{u}(t))$ such that
\begin{equation}
\label{eq:robot_eqn}
\begin{gathered}
\bm{u}^*(t)=\operatorname*{arg\,min}_{\bm{u}} \Phi(\bm{q}(t),\bm{u}(t),t),   \\  
\mbox{subject to } \dot{\bm{q}}(t)=g(\bm{q}(t),\bm{u}(t))
\\ \left\|\bm{u}(t)\right\|\leq u_{max} ,
\end{gathered}
\end{equation}
where $\bm{q}\in Q$ is the state space and $\bm{u}\in U$ denotes the set of controls. 

Mathew and Mezi{\'c}~\cite{mathew2011metrics} mainly consider first-order, \mbox{$\dot{\bm{q}}(t)=\bm{u}(t)$}, and second-order systems, \mbox {$\ddot{\bm{q}}(t)=\bm{u}(t)$}. They derive closed-form solutions that approximate the solution to Eq.~(\ref{eq:robot_eqn}) by discretizing the exploration time and solving for the optimal control input  that maximizes the rate of decrease of Eq.~(\ref{eq:robot_eqn}) at each time-step. Miller and Murphey~\cite{miller2013} use a projection-based trajectory optimization method that solves a first-order approximation of Eq.~(\ref{eq:metric_ergodicity}) at each iteration using linear quadratic regulator techniques~\cite{hauser2002projection}.  

\subsection{Cross-Entropy Trajectory Optimization}
\label{sec:CEM}
Consider a robot whose dynamics is described by the function  $g\colon Q \times U \to TX $, such that
\begin{equation}
\label{eq:dynamics}
\dot{\bm{q}}(t)=g(\bm{q}(t),\bm{u}(t)) 
\end{equation}
In addition, the robot is subject to constraints such as actuator bounds and obstacles in the environment. These constraints are expressed through the vector of inequalities
\begin{equation}
\label{eq:constraints}
F(\bm{q}(t))\geq 0,
\end{equation}
The goal is to compute the optimal controls $\bm{u}^*$ over a time horizon $t \in (0,t_f]$ that  minimize a cost function such that
\begin{equation}
\label{eq:CEM}
\begin{gathered}
\bm{u}^*(t)=\operatorname*{arg\,min}_{\bm{u}}\int_0^{t_f}  C(\bm{u}(t),\bm{q}(t))dt,   \\  
\mbox{subject to } \dot{\bm{q}}(t)=g(\bm{q}(t),\bm{u}(t)),\\
 F(\bm{q}(t))\geq 0 , \\
\bm{q}(0)=\bm{q}_0 ,
\end{gathered}
\end{equation}
where $\bm{q_0}$ is the initial state af the robot and $C\colon U \times Q \to \mathbb{R} $ is a given cost function.  

\subsubsection{Trajectory Parameterization}
Following the notation in~\cite{kobilarov2012cross},
a trajectory defined by the controls and states over the time
interval $[0,T]$ is denoted by the function \mbox {$\pi \colon [0,T] \to U \times Q$}, i.e. $\pi(t)=( \bm{u}(t) , \bm{q}(t))$ for all $t \in [0, T].$ The space of all trajectories originating at $\bm{q}_0$ and satisfying Eq.~(\ref{eq:dynamics}) is given by
\begin{equation}
\begin{aligned}
P=\{ \pi \colon t \in [0, T] \to \left( \bm{u}(t),\bm{q}(t)\right)| 
\dot{\bm{q}}(t)=g(\bm{q}(t),\bm{u}(t)), \\
 \bm{q}(0)=\bm{q}_0, T>0 .\} 
 \end{aligned}
\end{equation}
Let us consider a finite-dimensional parameterization of trajectories in terms of vectors $\bm{z} \in Z$ where $Z\subset \mathbb{R}^{n_z}$ is the parameter space. Assuming that the parameterization is given by a function $\varphi:Z \to P$ according to
\begin{equation}
\pi=\varphi(z). 
\end{equation}
The $(\bm{u},\bm{q})$ tuples along a trajectory parameterized by $\bm{z}$ are written as $\pi(t)=\varphi(\bm{z},t)$. One choice of parameterization is to use motion primitives defined as \mbox{$\bm{z} = (\bm{u}_1, \tau_1,...,\bm{u}_j, \tau_j)$} where each $u_i$, for $1 \leq i \leq j$, is a constant control input applied for duration $\tau_i$. For differentially flat systems~\cite{murray1995differential}, a sequence of states, $\bm{z} =(\bm{q}_1,...,  \bm{q}_j)$ can also be used. Another option is to use RRT or probabilistic roadmaps (PRM) to sample trajectories as demonstrated in~\cite{kobilarov2012cross}.

Now, we can define a cost function, $J \colon Z \to \mathbb{R}$ , in terms of the trajectory parameters as
\begin{equation}
\label{eq:CEM_costfunc}
\begin{gathered}
 J(\bm{z})=\int_0^T C(\varphi(\bm{z},t)) dt
\end{gathered}
\end{equation}
Eq.~(\ref{eq:CEM}) can be restated as finding the optimal $(\bm{u}^*,\bm{q}^*)=\varphi(\bm{z}^*)$ such that
\begin{equation}
\label{eq:CEM_cost}
\bm{z}^*= \operatorname*{arg\,min}_{\bm{z} \in Z_{con}} J(\bm{z}).
\end{equation}
where the constrained parameter space $Z_{con} \subset Z$ is
the set of parameters that satisfy the boundary conditions and
constraints in Eq.~(\ref{eq:CEM}). 

\subsubsection{Cross-Entropy Method}
In this work, we employ the cross entropy (CE) method to optimize the parameters of the trajectory. There are other sampling-based global optimization methods such as Bayesian optimization~\cite{snoek2012practical}, simulated annealing~\cite{van1987simulated}, and other variants of stochastic optimization~\cite{zhigljavsky2007stochastic} that can also be used to optimize parameterized trajectories. We use the CE method because it  utilizes importance sampling to efficiently compute trajectories that have lower costs after few iterations of the algorithm, and it has been shown to perform well for trajectory optimization of nonlinear dynamic systems~\cite{kobilarov2012cross}.  

The CE method treats the optimization in Eq.~(\ref{eq:CEM_cost}) as an estimation problem of rare-event probabilities. The rare event of interest in this work is to find a parameter $z$ whose cost $J(z)$ is very close to the cost of an optimal parameter $z^*$. It is assumed that the parameter $z \in Z$ is sampled from a Gaussian mixture model defined as
\begin{equation}\label{eq:GMM}
p(z;v)=\sum_{k=1}^{K}\frac{w_k}{\sqrt{(2\pi)^{n_z} |\Sigma_k|}} e^{-\frac{1}{2}(z-\mu_k)^T \sigma_k^{-1}(z-\mu_k)}
\end{equation}
where $v=(\mu_1,\Sigma_1,...,\mu_K,\Sigma_K,w_1,...,w_K)$ corresponds to $K$ mixture components with means $\mu_k$, covariance matrices $\Sigma_k$, and weights $w_k$, where $\sum_{k=1}^{K}w_k=1$. 

The CE method involves an iterative procedure where each iteration has two steps: \textit{(i)} select a set of parameterized trajectories from $p(z;v)$ using importance sampling~\cite{rubinstein2013cross} and evaluate the cost function $J(z)$, (ii) use a subset of \textit{elite} trajectories\footnote{A fraction of the sampled trajectories with the best costs form an elite set. See~\cite{rubinstein2013cross} for details.} and update $v$ using expectation maximization~\cite{mclachlan2004finite}. After a finite number of iterations $p(z;v)$ approaches to a delta distribution, thus the sampled trajectories remains unchanged. For implementation details, the reader is referred to~\cite{de2005tutorial}.

 \begin{figure*}[bt]
 	\centering
 	\includegraphics[width=0.9\linewidth]{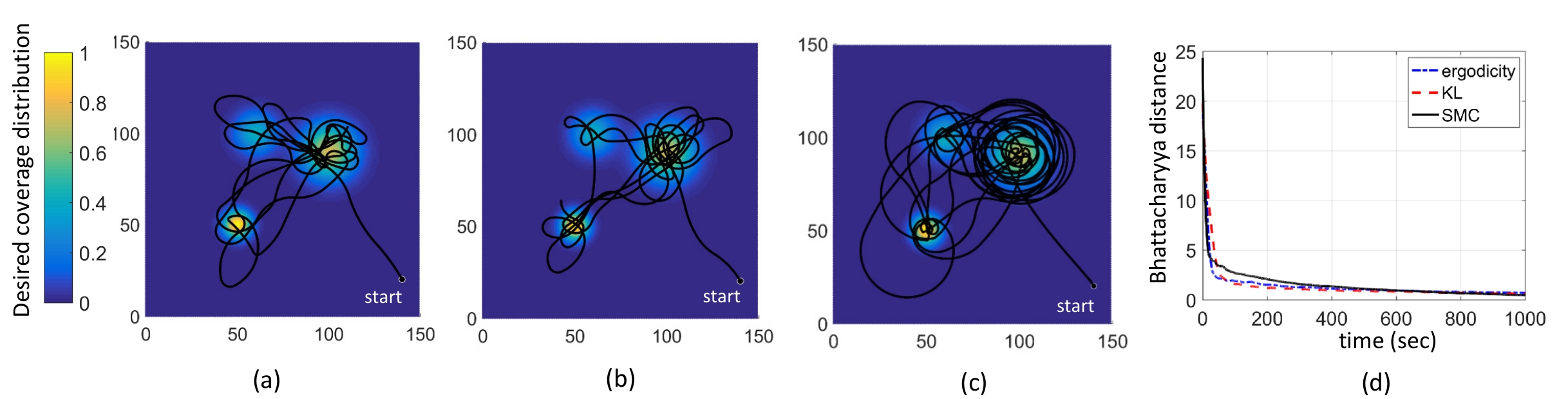}
 	\caption{Ergodic coverage results for three different implementation: (a) Ergodic-STOEC: minimizing Eq.(\ref{eq:metric_ergodicity}), (b) KL-STOEC: minimizing Eq.(\ref{eq:KLdiv}),  (c) SMC implementation in Mathew \textit{et. al}~\cite{mathew2013experimental}. Ergodic coverage performance is assessed through plotting the Bhattacharyya distance  between the coverage distribution $\xi(\bm{x})$ and the time-average statistics distribution $\Gamma(\bm{x})$ in (d) for the different implementations. Notice that (a) and (b) have similar decay rates, while the decay rate of (c) is smaller.}
 	\label{fig:Comparison}
 \end{figure*}

\begin{figure}[htb]
	\centering
	\includegraphics[width=.6\linewidth]{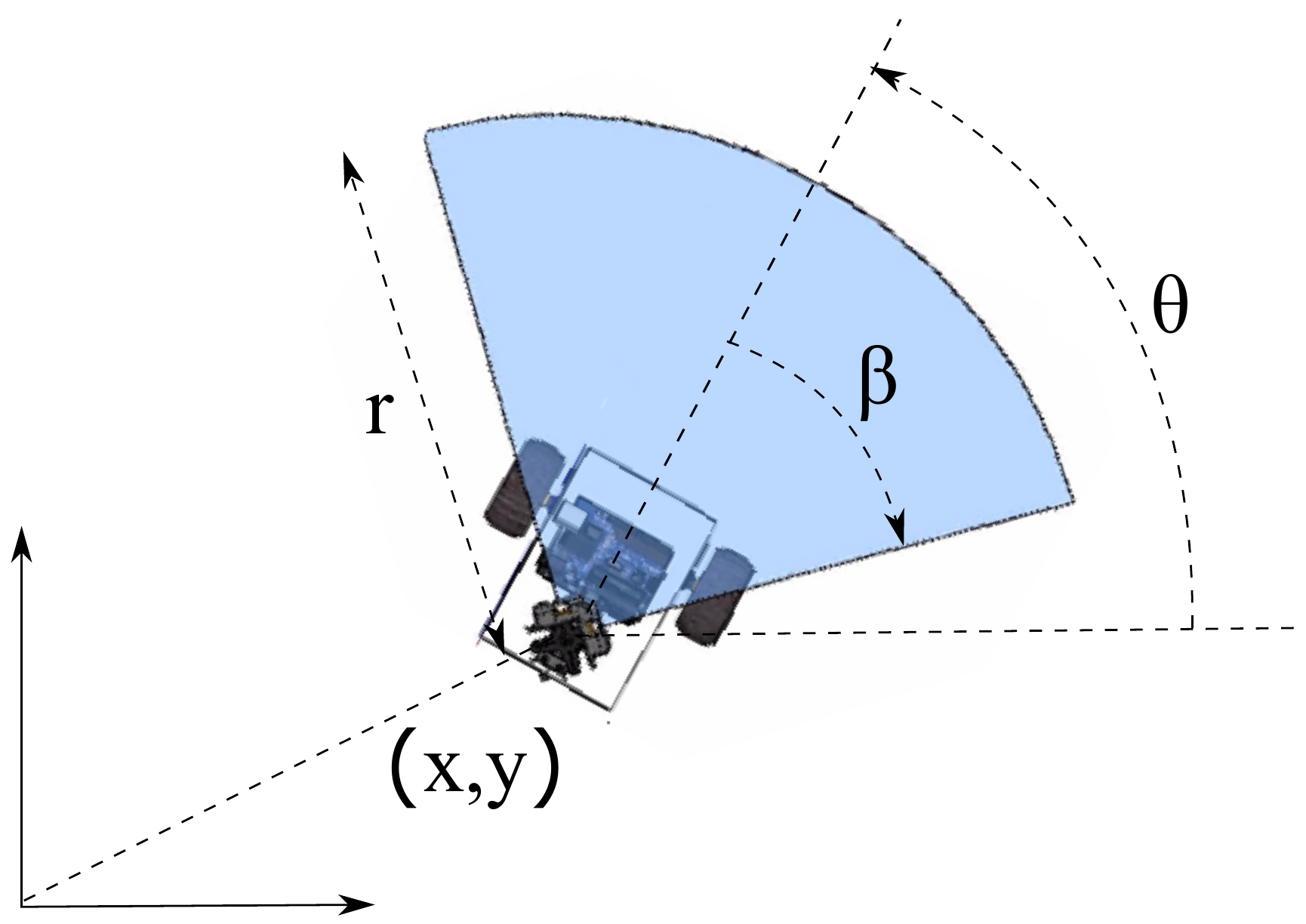}
	\caption{Forward looking beam sensor model (e.g. ultrasonic sensor): an example of a sensor footprint of the robot. The variables $(x, y, \theta)$ represent the state of the robot, whereas $r$ and $\beta$ represent the bearing measurement limits of the sensor.}
	\label{fig:beamsensor}
\end{figure}
\section{Stochastic Trajectory Optimization for Ergodic Coverage (STOEC)}
\label{sec:Stochastic_Ergodic}
In this section, we introduce a formulation that optimizes an ergodic coverage objective using the stochastic trajectory optimization algorithm presented in Section~\ref{sec:CEM}. In this formulation, we pose ergodic coverage as a constrained optimization problem subject to constraints associated with obstacle avoidance and with the motion model of the robot. We start by redefining the time average statistics of a robot's trajectory to capture the robot's sensor footprint. We then demonstrate that obstacle avoidance can be easily incorporated into this framework by giving penalties to the sampled trajectories that collide with arbitrarily-shaped obstacles or that cross restricted regions.

Let us start with two key observations. First,  Eq.~(\ref{eq:time-average}) assumes that a robot's trajectory is composed of Dirac delta functions. In other words, it is assumed that the sensor  footprint is negligible compared to the size of the domain. This assumption is adopted to efficiently calculate the Fourier coefficients of the distributions given by Eq.~(\ref{eq:Fourier_coefficients}). In many practical applications, the robots can be equipped with sensors of various types  (camera, lidar, ultrasonic, etc.), which are hardly captured by the Dirac delta function. Having control over the sensor footprint opens up interesting and useful strategies for coverage:  a robot equipped with a wide sensor footprint should be able to perform a coarse ergodic coverage of the domain compared to a robot equipped with a narrow sensor footprint. Similarly, the optimal ergodic-coverage trajectory of a ground vehicle equipped with a forward looking ultrasonic sensor would be different than an aerial vehicle equipped with a downward-looking sensor.

Second, as also stated in ~\cite{miller2013},  the motivation in using the norm on the Fourier coefficients is to obtain an objective function that is differentiable with respect to the trajectory. When a sampling-based approach is used for trajectory optimization, there is no need for a differentiable objective function. There are many other measures used in information theory and statistics to compare two distributions. Here, we consider Kullback-Leibler (KL) divergence, which measures the relative entropy between two distributions,
\begin{equation}
\label{eq:KLdiv}
D_{KL}(\Gamma \vert \xi)=\int_{X}\Gamma(\bm{x}) \log\left(\frac{\Gamma(\bm{x})}{\xi(\bm{x})}\right)d\bm{x}.
\end{equation}
KL divergence can encode an ergodic coverage objective without resorting to the spectral decomposition of the desired coverage distribution, whose accuracy is limited by the number of basis functions used. Additionally, we can explicitly measure the ergodicity of a trajectory without resorting to simplifications in the representation of a robot's sensor footprint. 

In our formulation, the optimal control problem posed in Eq. (\ref{eq:CEM_cost}) can be solved by defining the cost function $J(\textbf{z})$ to be 
\begin{equation}
J(\textbf{z}) = D_{KL}(\Gamma \vert \xi)
\end{equation}
where $z$ is sampled from a Gaussian mixture model defined in Eq. (\ref{eq:GMM}).

\subsection{Sensor Footprints}
In this paper, we consider two sensor footprints: (1) Gaussian sensor footprint, and (2) Forward-looking beam sensor footprint. Assuming there are $N$ robots, let $f_j(.)$ be the sensor footprint of the $j$-th robot. The time-average statistic, $\Gamma(\bm{x})$, can be defined as,
 \begin{equation}\label{eq:general time-average}
 \Gamma(\bm{x}) = \eta\sum_{j=1}^{N}\int_{0}^{t} f_j(\bm{x}-\gamma_j(\tau)) d\tau ,
 \end{equation}
 where $\gamma_j$ is the trajectory of the $j$-th robot, and $\eta$ is a normalizing factor such that $\Gamma(\bm{x})$ is a pdf. A Gaussian sensor footprint is described as,
 \small
\begin{equation}
f_j(\bm{x}-\gamma_j(\tau)) = \frac{1}{\sqrt{(2\pi)^{d}|\Sigma| }}\exp(-\frac{1}{2}(\bm{x}-\gamma_j(\tau))^T\Sigma^{-1}(\bm{x}-\gamma_j(\tau)))
\end{equation}
\normalsize
whereas the sensor footprint of a beam model is  defined by a radius, $r$, and angle of view, $\beta$, similar to an ultrasonic sensor as shown in Fig.~\ref{fig:beamsensor}. The sensor footprint of the beam model is parameterized as
\begin{equation}
\small
f_j(\bm{x}-\gamma_j(\tau)) = \left\{
\begin{array}{ll}
1, & \mbox{if } |\bm{x}-\gamma_j(\tau)| \leq r ~\mbox{and}~ \\&~~~\theta_\tau - \frac{\beta}{2}\leq \arctan(\frac{(\bm{x}-\gamma_j(\tau))_2}{(\bm{x}-\gamma_j(\tau))_1})\leq \theta_\tau + \frac{\beta}{2} \\
0, & \mbox{otherwise}. 
\end{array}
\right.
\normalsize
\end{equation}
where $\theta_\tau$ is the heading of the robot at time $\tau$, $(\bm{x}-\gamma_j(\tau))_i$ is the $i$-th element of the vector $(\bm{x}-\gamma_j(\tau))$.

\subsection{Obstacle Avoidance}
Suppose that the domain $X$ contains $p$ obstacles denoted by $O_1,..., O_p \subset X$. We assume that the robot at state $\bm{q}$ is occupying a region $A(\bm{q})\subset X$. Borrowing the notation in \cite{kobilarov2012cross}, let the function \textbf{prox}($A_1,A_2$) return the closest Euclidean distance between two sets $A_{1,2} \subset X$. This function returns a negative value if the two sets intersect. Therefore, for an agent to avoid the obstacles $O_1,...,O_p$, we impose a constraint of the form of Eq.~(\ref{eq:constraints}) expressed as,
\begin{equation}\label{eq:constraint}
F(\bm{q}(t)) = \min_i \mbox{\textbf{prox}}(A(\bm{q}(t)),O_i), ~ \forall t\in[0,\infty).
\end{equation}

The sampled trajectories that does not  satisfy the constraints of Eq. (\ref{eq:constraints})  are penalized.

\subsection{Example: Dubins Car}
The trajectory optimization framework can be applied to any robot whose dynamics are given as (\ref{eq:dynamics}) given that we can parameterize the space of all trajectories satisfying (\ref{eq:dynamics}) using motion primitives. We will consider the Dubins car model whose motion is restricted to a plane. The state space for such a model is $Q = SE(2)$ with state $\bm{q} = (x, y , \theta)$, where $(x, y)$ are the Cartesian coordinates of the robot, and $\theta$ is the orientation of the robot in the plane. The dynamics of a dubins car is defined by,
\begin{equation}\label{dubins}
\dot{x} = u\cos\theta ,~~~  \dot{y} = u\sin\theta,~~~   \dot{\theta} = v
\end{equation}
where $v \in [v_{min}, v_{max}]$ is a controlled forward velocity, and $w \in [w_{min}, w_{max}]$ is a controlled turning rate.

We can represent a trajectory satisfying (\ref{dubins}) as a set of connected motion primitives consisting of either straight lines with constant velocity $v$ or arcs of radius $v/w$. We define a primitive by a constant controls ($v, w$). The duration of each primitive is constant and $\tau > 0$. We parameterize the trajectory of the robot using $m$ primitives, and this finite dimensional parameterization is represented by  a vector $\bm{z} \in \mathbb{R}^{2m}$ such that,
\begin{equation}
\bm{z} = (v_1,w_1, ..., v_m, w_m) 
\end{equation}

The $j^{th}$ primitive ends at time $t_j = j\tau$ for $j \in {1,..,m}$. For any time $t \in [t_j, t_{j+1}]$, the parameterization $\varphi_z = (v, w, x, y ,\theta)$ is given by,
\begin{align}\nonumber
v(t) &= v_{j+1}
\\\nonumber
w(t) &= w_{j+1}
\\
\theta(t) &= \theta_j + w_{j+1}\Delta t_j
\\\nonumber
x(t) &= \left\{
\begin{array}{ll}
x(t_j) + \frac{v_{j+1}}{w_{j+1}}(\sin(\theta(t))-\sin\theta_j),  & \mbox{if } w_{j+1}\neq0 \\
x(t_j) + v_{j+1} \Delta t_j\cos\theta_j, &  otherwise
\end{array}
\right.
\\\nonumber
y(t) &= \left\{
\begin{array}{ll}
y(t_j) + \frac{v_{j+1}}{w_{j+1}}(\cos\theta_j - \cos(\theta(t))),  & \mbox{if } w_{j+1}\neq0 \\
y(t_j) + v_{j+1} \Delta t_j\sin\theta_j, &  otherwise
\end{array}
\right.
\end{align}
where $\theta_j = \theta(t_j)$ and $\Delta t_j = t - t_j$.

\section{Results}
\label{sec:Results}
In this section, we first compare our algorithm with the implementation of Mathew \textit{et. al} ~\cite{mathew2013experimental} in an unconstrained environment. In Section~\ref{sec:Heterogeneous}, we  demonstrate that our formulation can capture coordination of multiple robots equipped with different sensing capabilities. It is also straightforward to encode additional tasks besides ergodic coverage. In Section~\ref{sec:Point2Point}, we present an example in which a robot is directed to reach a destination in a desired time while performing ergodic coverage.

\subsection{Comparisons}
\label{sec:Comparisons}
We first compare the coverage performance of three different implementations of the ergodic coverage algorithm: (i) We generate a trajectory that minimizes the rate of change of the metric for ergodicity given by Eq.~(\ref{eq:metric_ergodicity}) as presented in Mathew \textit{et. al}~\cite{mathew2013experimental}, which will be referred to as spectral-multiscale coverage (SMC) algorithm. (ii) We use STOEC to optimize the trajectory such that Eq.~(\ref{eq:metric_ergodicity}) is minimized. This implementation will be referred as Ergodic-STOEC. (iii) We use STOEC to optimize the trajectory such that Eq.~(\ref{eq:KLdiv}) is minimized. This will be referred as KL-STOEC.

We perform numerical simulations for each of the aforementioned implementations. 
We define $\xi(\bm{x})$, as a Gaussian mixture model\footnote{There are no restrictions on the choice of representation for $\xi(\bm{x})$.} as shown in Fig.~\ref{fig:Comparison}. For visualization, $\xi(\bm{x})$ is normalized such that the maximum value of the desired coverage distribution takes the value of 1. The initial position of the robot is randomly selected. The robot is assumed to have bounded inputs $v \in [0.1, 5]$ m/s and $w \in [-0.2, 0.2] $ rad/s. For SMC, the total simulation time is $T=$ 1000~sec, and the time step is 0.1~sec. For STOEC, we run 20 stages of the algorithm using 5 primitives with a receding horizon of \mbox{$T$= 50} sec. A single stage of the algorithm corresponds to generating a trajectory over a finite horizon of $T$ seconds by optimizing Eq.~(\ref{eq:CEM_costfunc}). Therefore, the full trajectory is 1000 sec. STOEC is seeded with 40 sampled-trajectories.
 
The results are shown in Fig.~\ref{fig:Comparison}. Since the metric for ergodicity and KL divergence both measure similarity between two distributions, we need to choose another metric to compare $\Gamma(\bm{x})$ of the trajectory optimized by KL-STOEC and $\xi(\bm{x})$, and to compare $\Gamma(\bm{x})$ of the trajectory optimized by Ergodic-STOEC and $\xi(\bm{x})$. To assess the coverage performance of the algorithms, we use Bhattacharyya distance that measures the similarity of two probability distributions.  The Bhattacharyya distance is measured as,
 \begin{equation}
 D_B(\Gamma , \xi)=-\ln\left(\int_X \sqrt{\Gamma(\bm{x}) \xi(\bm{x})} d\bm{x}\right).
 \end{equation}
Table 1 shows the computation time\footnote{The code runs on MATLAB 2016a in Windows 10 on a laptop with i7 CPU and 8 GB RAM.} for each algorithm for the example in Fig.\ref{fig:Comparison}. We should mention that the computational time of STOEC varies depending on the number of trajectory samples that are seeded to the algorithm and the number of primitives. With around 40 sampled trajectories, the algorithm converges to a solution more quickly than SMC. It is also important to note that STOEC plans over a finite time horizon as opposed to the greedy approach in SMC. The metric for ergodicity, given by Eq.~(\ref{eq:metric_ergodicity}), gives higher weights to the low frequency components. Therefore, the algorithm favors large scale exploration of the coverage distribution.  In the accompanying video of Fig.~\ref{fig:Comparison}, it can be noticed that the metric for ergodicity given by Eq.~(\ref{eq:metric_ergodicity}) results in a trajectory that frequently crosses from one region to another, whereas Eq.~(\ref{eq:KLdiv}) results in a coverage behavior that gradually expands from one region to another.

\begin{table}
	\label{table:runtime}
	\centering
	\caption{Comparison of the computation time}
	\begin{tabular}{lccc}
		\hline
		Method        & \begin{tabular}[c]{@{}c@{}}Computing \\ 1 stage (50 sec) \\ \end{tabular} & \begin{tabular}[c]{@{}c@{}}Computing the  \\full trajectory (1000 sec)\end{tabular} \\ \hline \\
		\textbf{KL-STOEC}        & \textbf{0.5}    &\textbf{10.0} \\
		Ergodic-STOEC   & 1.2    & 24.0 \\
		SMC   & 1.7    & 35.0 \\
	\end{tabular}
\end{table}

\begin{figure}[tbh]
	\centering
	\includegraphics[width=\columnwidth]{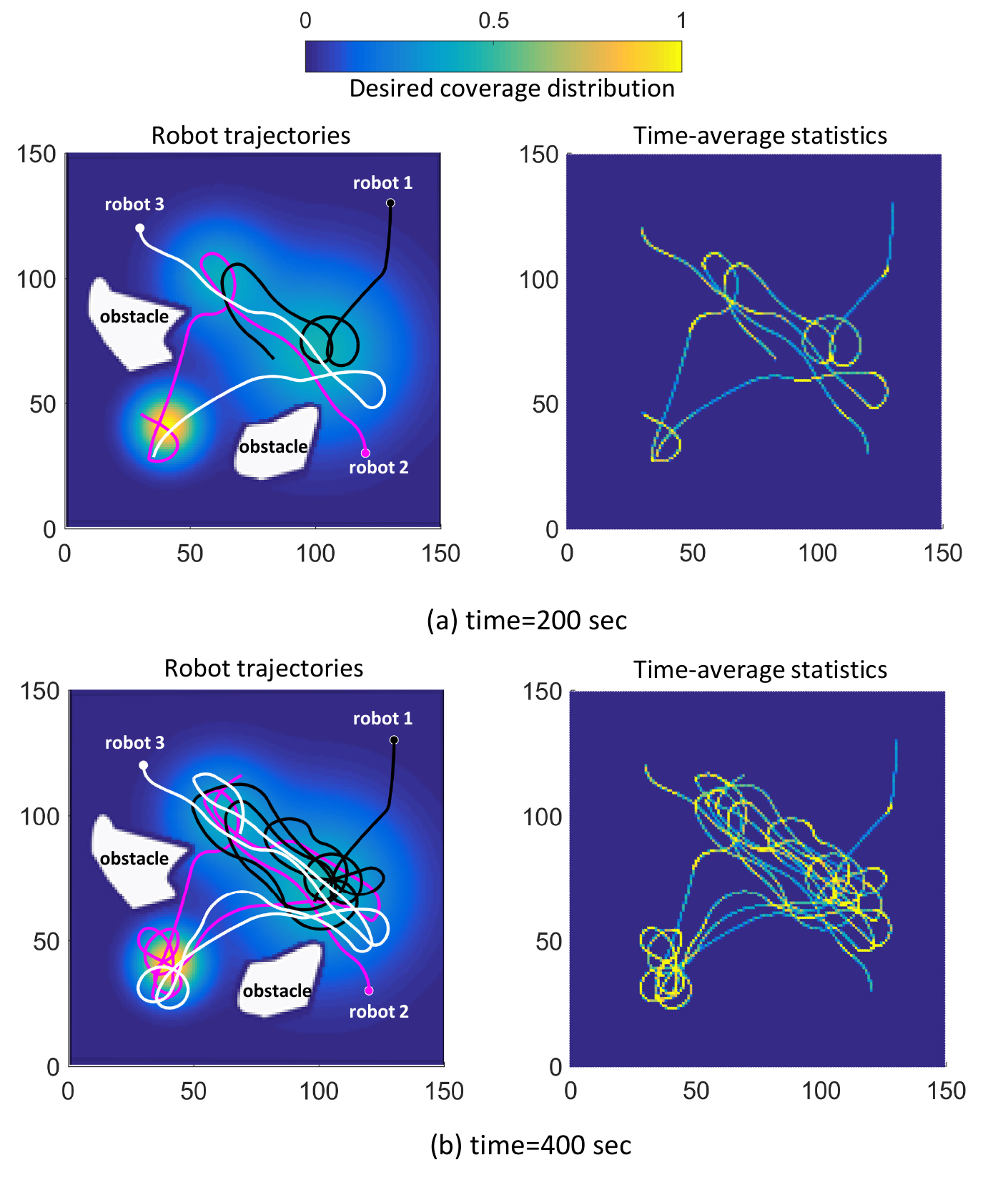}
	\caption{Ergodic coverage results for three robots with Dirac delta sensor footprint. The left figures show the robots trajectories, and the right figures show the corresponding time-average statistics. The results are shown for two different instants of the KL-STOEC.}
	\label{fig:Small_sensor}
\end{figure}

\begin{figure}[bth]
	\centering
	\includegraphics[width=\columnwidth]{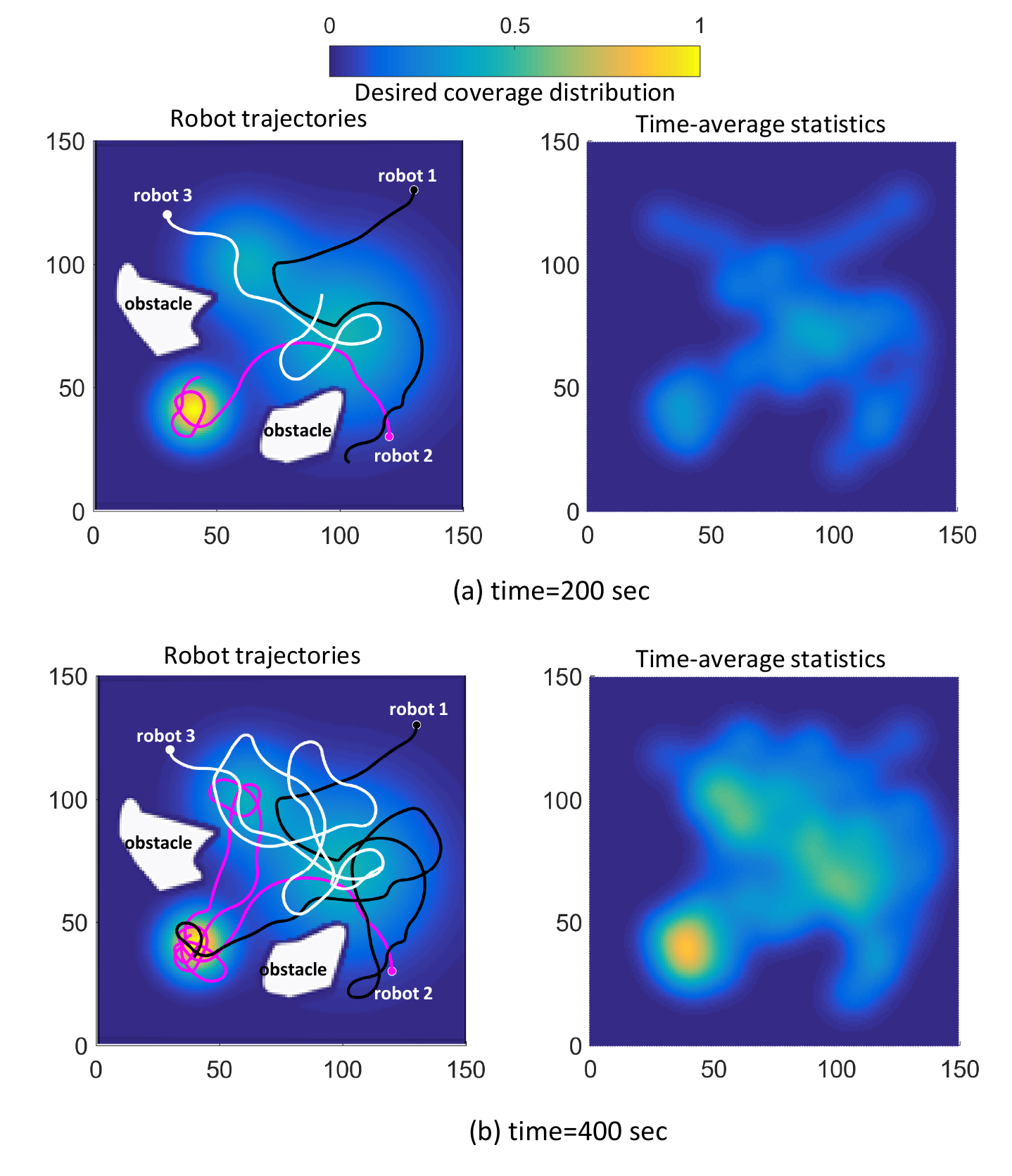}
	\caption{Ergodic coverage results for three robots with wide Gaussian sensor footprint. The left figures show the robots trajectories, and the right figures show the corresponding time-average statistics. The results are shown for two different instants of the KL-STOEC.Large variance Gaussian sensor footprint}
	\label{fig:Large_sensor}
\end{figure}
\subsection{Sensor Footprints}
\label{sec:Heterogeneous}
In this section, we demonstrate that our framework can capture multi-agent systems with different sensing capabilities. It is important to note that the formulation of the  ergodic coverage algorithm assumes that robots have perfect communication, thus have access to the trajectory plans of other robots. The robot trajectories are sequentially optimized by taking into account the time-average statistics of every robot's trajectory. Trajectory optimization also assumes deterministic motion models.

Fig.~\ref{fig:Small_sensor} shows the ergodic coverage of a desired distribution using KL-STOEC assuming a Dirac delta sensor footprint. We run 10 stages of the algorithm using 5 primitives with a receding horizon of \mbox{$T$ = 40 sec}. The robot is assumed to have bounded inputs $v \in [0.1, 5]$ m/s and $w \in [-0.2, 0.2] $ rad/s. Although the coverage focuses on the regions where the measure of the desired coverage distribution, $\xi$, is greater than 0, the algorithm needs a much longer time for time-average statistics to converge to the desired distribution. Fig.~\ref{fig:Large_sensor} demonstrates results with a wide Gaussian sensor footprint using the same parameters. Fig.\ref{fig:Coverage_error} shows the errors measured using the Bhattacharyya distance. The computational time for these two scenarios in Fig.~\ref{fig:Small_sensor} and Fig.~\ref{fig:Large_sensor} are 0.7 sec/stage and 1.7 sec/stage, respectively.
Fig.~\ref{fig:Hetero_sensor} shows another example where there are three robots equipped with different sensing capabilities. The sensor footprint of each robot is shown in {\mbox{Fig.~\ref{fig:Hetero_sensor}~(c)}. 

Note that, in these examples, the robots can continue to perform ergodic coverage of the domain indefinitely. That is, once the time-average statistics converge to the desired distribution, any new motion will result in a difference between the two distributions, $\Gamma(\bm{x})$ and $\xi(\bm{x})$. Thus, the robots will continue covering the domain such that time average-statistics again converges to the desired distribution (see the accompanying video). This is a very useful trait for surveillance applications, where the robots need to continuously monitor a region while spending more time in regions that have high values marked by the desired coverage distributions.

\begin{figure}[bt]
	\centering
	\includegraphics[width=0.7\columnwidth]{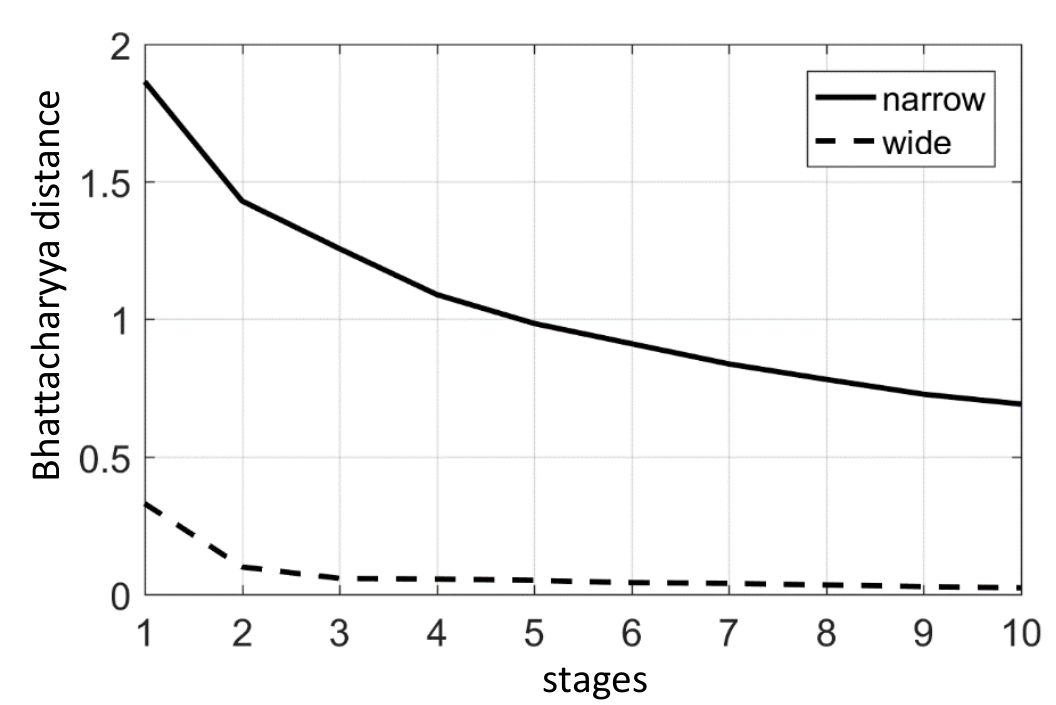}
	\caption{Ergodic coverage performance comparison for the robots with narrow vs. wide Gaussian footprints.}
	\label{fig:Coverage_error}
\end{figure}


\begin{figure}[tbh]
	\centering
	\includegraphics[width=\columnwidth]{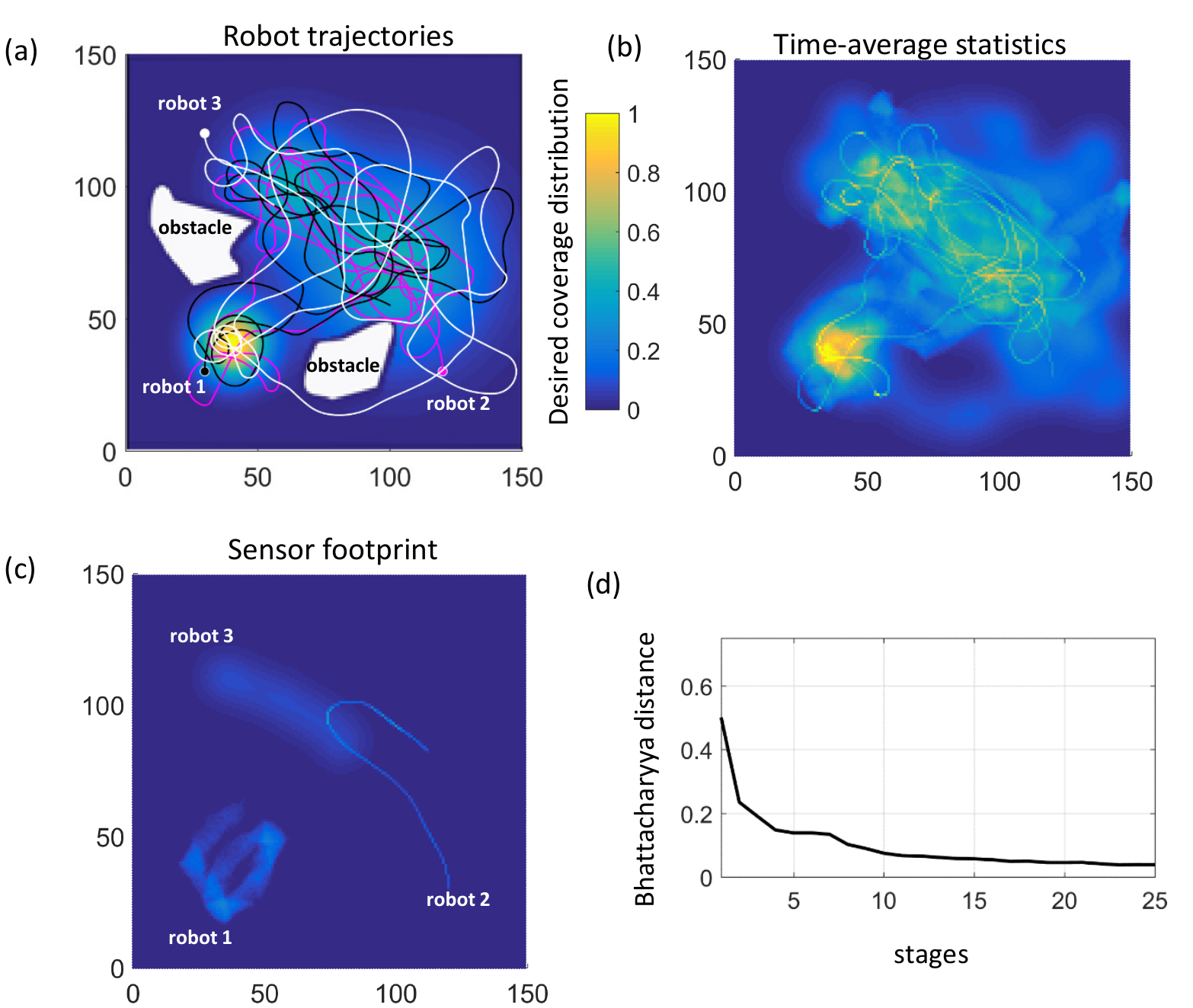}
	\caption{Ergodic coverage results for three robots equipped with a forward-looking beam sensor footprint, a Dirac delta sensor footprint, and a wide Gaussian sensor footprint, respectively. The resulting trajectories of the robots using KL-STOEC implementation are shown in (a) for 25 concatenated $T = 40$~sec segments. The  time-average statistics is shown in (b). The different footprints of the three robots are shown in (c). Finally, (d) shows the Bhattacharyya distance between the coverage distribution $\xi(\bm{x})$ and the time-average statistics distribution $\Gamma(\bm{x})$ as a function of the number of stages.}
	\label{fig:Hetero_sensor}
\end{figure}

\begin{figure}[bth]
	\centering
	\includegraphics[width=0.5\columnwidth]{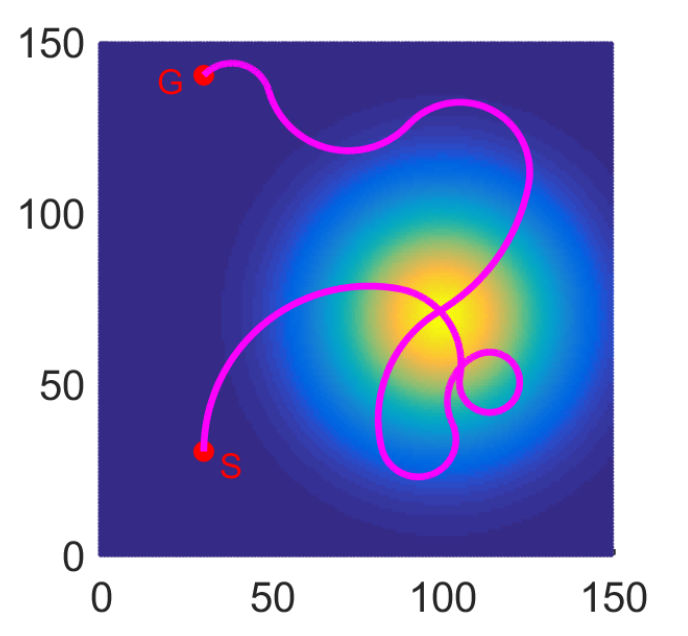}
	\caption{The robot starts at \textbf{S}:start and is supposed to reach point \textbf{G}:goal in 500 seconds using a maximum of 10 primitives.}
	\label{fig:Point2Point}
\end{figure}

\subsection{Point-To-Point Planning}
\label{sec:Point2Point}
Our algorithm can be extended to encode additional tasks beside ergodic coverage. We present an example in which the robot reaches a desired state in a given time while performing ergodic coverage along the way. This task is easily performed by adding a term to the objective function, Eq.~(\ref{eq:KLdiv}), that penalizes the error between the state of the robot at the end of the horizon, $\bm{q}(T)$= 500 sec, and the desired state of the robot $\bm{q}_{d}$. Thus, the cost function is defined as,
\begin{equation}
C(\bm{z},\bm{q}) =  D_{KL}(\Gamma \vert \xi) + \alpha\|\bm{q}(T) - \bm{q}_{d}  \|_2.
\end{equation}
where $\alpha$ is a weighting parameter, chosen heuristically, which specifies how \textit{hard} the constraint of reaching a desired state at time $T$ is. As $\alpha$ increases, the robot is guaranteed to reach its final desired state. We simulate an example of such a scenario in Fig.~\ref{fig:Point2Point}. The robot is assumed to have bounded inputs $v \in [0.1, 5]$ m/s and $w \in [-0.2, 0.2] $ rad/s. How to systematically choose the parameter $\alpha$ is an open question and is left as a future work.

\section{Conclusions}
This work extends the ergodic coverage algorithm presented in~\cite{mathew2011metrics} to robots operating in constrained environments. The sampling-based trajectory optimization presented in this work allows obstacle  avoidance by penalizing sampled trajectories that collide with arbitrarily-shaped obstacles or that cross restricted regions. We demonstrated that  Kullback-Leibler divergence can also be used to encode an ergodic coverage objective without resorting to spectral decomposition of the desired coverage distribution, whose accuracy is limited by the number of basis functions used. Compared to~\cite{mathew2011metrics},  which is formulated as a first-order iterative optimization, our formulation allows for trajectory optimization over a receding horizon.  We believe, our formulation will be of interest to the wider robotics community because it captures sensor footprints, avoids obstacles, and can be applied to nonlinear dynamic systems~\cite{kobilarov2012cross}. Future work will focus on decentralization of the ergodic coverage algorithm to address multi-agent systems with limited communication.





\section*{ACKNOWLEDGMENT}
The authors would like to thank Marin Kobilarov for useful discussions and assistance in the implementation of the cross entropy method.

\bibliographystyle{IEEEtran}
\bibliography{references}
\end{document}